\documentclass{preprint}

\usepackage{graphicx}%
\usepackage{xcolor}%
\usepackage{booktabs}%
\usepackage{spverbatim}
\usepackage{tabularx}
\usepackage{longtable}
\usepackage{subcaption}
\usepackage{colortbl}
\usepackage{makecell}
\usepackage{hyperref}
\usepackage{mdframed}
\usepackage{enumitem}
\usepackage[numbers,comma,sort&compress]{natbib} 

\newenvironment{prompt}
{\begin{mdframed}[innertopmargin=0pt]%
\setlength{\parindent}{0pt}%
\setlength{\parskip}{1em}%
\footnotesize\ttfamily\hyphenchar\font=`\-\spaceskip=.5em plus .5em\xspaceskip=.5em%
}
{%
\par%
\end{mdframed}%
}

\title{A Multi-Stage Large Language Model Framework for Extracting Suicide-Related Social Determinants of Health}

\author[1]{Song Wang}
\author[2]{Yishu Wei}
\author[2]{Haotian Ma}
\author[2]{Max Lovitt}
\author[2]{Kelly Deng}
\author[2]{Yuan Meng}
\author[2]{Zihan Xu}
\author[2]{Jingze Zhang}
\author[2]{Yunyu Xiao}
\author[3]{Ying Ding}
\author[4]{Xuhai Xu}
\author[1]{Joydeep Ghosh}
\author[2,*]{Yifan Peng}

\affil[1]{Cockrell School of Engineering, The University of Texas at Austin, Austin, Texas, USA}
\affil[2]{Population Health Sciences, Weill Cornell Medicine, New York, New York, USA}
\affil[3]{School of Information, The University of Texas at Austin, Austin, Texas, USA}
\affil[4]{Department of Biomedical Informatics, Columbia University, New York, New York, USA}
\affil[*]{Corresponding author(s). Email(s): \url{yip4002@med.cornell.edu}}

\begin{document}

\maketitle

\begin{abstract}

\noindent
\textbf{Background:}
Understanding social determinants of health (SDoH) factors contributing to suicide incidents is crucial for early intervention and prevention. However, data-driven approaches to this goal face challenges such as long-tailed factor distributions, analyzing pivotal stressors preceding suicide incidents, and limited model explainability.

\textbf{Methods:}
We present a multi-stage large language model framework to enhance SDoH factor extraction from unstructured text. Our approach was compared to other state-of-the-art language models (i.e., pre-trained BioBERT and GPT-3.5-turbo) and reasoning models (i.e., DeepSeek-R1). We also evaluated how the model's explanations help people annotate SDoH factors more quickly and accurately. The analysis included both automated comparisons and a pilot user study.

\textbf{Results:}
We show that our proposed framework demonstrated performance boosts in the overarching task of extracting SDoH factors and in the finer-grained tasks of retrieving relevant context. Additionally, we show that fine-tuning a smaller, task-specific model achieves comparable or better performance with reduced inference costs. The multi-stage design not only enhances extraction but also provides intermediate explanations, improving model explainability.

\textbf{Conclusions:}
Our approach improves both the accuracy and transparency of extracting suicide-related SDoH from unstructured texts. These advancements have the potential to support early identification of individuals at risk and inform more effective prevention strategies.

\end{abstract}


\section{Introduction}\label{section:introduction}

Suicide remains a major global public health concern, presenting significant challenges for intervention and prevention. The underlying causes of suicidal thoughts and behaviors stem from a complex interplay of social, economic, political, and physical factors~\cite{who-2014-sdoh}, all of which fall under the Social Determinants of Health (SDoH) framework~\cite{hacker2022sdoh, liu2023suicide-sdoh}. A deep understanding of these factors is essential for developing targeted, evidence-based strategies for suicide prevention and early intervention. While there is growing interest in integrating suicide-related SDoH factors into structured electronic health records (EHRs), much of this information remains incomplete or inaccessible, as it is often embedded within unstructured text narratives~\cite{patra2021-sdoh-review}.

In recent years, advancements in natural language processing (NLP) have opened new opportunities for modeling, extracting, and analyzing information from clinical notes, including suicide-related SDoH factors~\cite{patra2021-sdoh-review,magoc2023nlp-sdoh}. However, three major challenges remain. First, suicide-related SDoH factors follow a long-tailed distribution, where a small subset of factors appear frequently while most are rare. Conventional NLP models, which excel at extracting common factors, often struggle to capture these rare yet critical ones~\cite{HAN2022-sdoh, wang2023sdoh, gabriel2024-bert-sdoh}. Second, understanding SDoH factors occurring preceding a suicide incident is crucial, as this period often represents a critical window where acute stressors and life events may directly influence an individual's decision~\cite{liu2023nvdrs,wang2023sdoh}. However, many models struggle with capturing temporal context~\cite{wang2023sdoh}. Finally, deep learning models are often criticized for their ``black-box'' nature~\cite{bai2021explainable,adadi2018explainable,hassija2023explainable,singh2024rethinkinginterpretabilityeralarge}, making it difficult to interpret their reasoning. This lack of explainability is particularly problematic in suicide studies, where trustworthiness and explainability are essential.

To address these challenges, we developed a multi-stage zero-shot large language model (LLM) framework for extracting suicide-related SDoH factors from unstructured narratives in a scalable and generalizable manner. This study focuses on SDoH factors occurring within the two weeks preceding suicide incidents. By leveraging zero-shot learning and self-explanation in LLMs~\cite{Guevara2024-sdoh}, our approach captures nuanced suicide-related SDoH factors from limited labeled data and generalizes effectively to new, unseen cases in complex clinical text~\cite{consoli2024-sdohgpt}. We evaluated our framework against state-of-the-art baselines (i.e., pre-trained BioBERT and GPT-3.5-turbo) and reasoning models (i.e., DeepSeek-R1), demonstrating its effectiveness. The multi-stage design also provides intermediate explanations of model decisions, enhancing interpretability and aiding in the identification of actionable insights for future interventions. A pilot user study further validated that experts using these explanations annotated suicide-related SDoH factors more quickly and accurately than through manual efforts alone. 

\section{Methods}\label{section:methods}

\subsection{Framework architecture}
\label{sec:framework}

\begin{figure}[tbp]
\centering
\includegraphics[width=.45\linewidth]{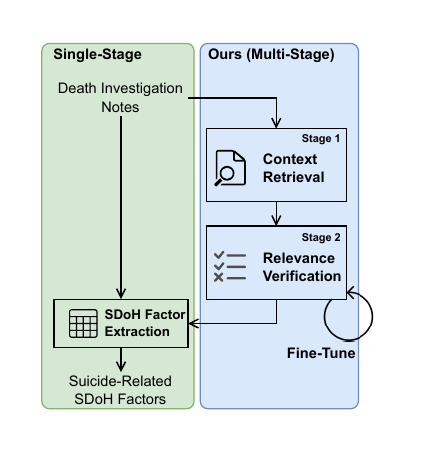}
\caption{Overview of the proposed multi-stage large language model framework for extracting suicide-related SDoH factors.}
\label{fig:overview}
\end{figure}

We present a multi-stage LLM framework designed to extract suicide-related SDoH factors from de-identified death investigation notes. As shown in Figure \ref{fig:overview}, the framework consists of two intermediate stages - context retrieval and relevance verification - and one final decision-making stage, which is SDoH factor extraction.

\subsubsection{Context retrieval}

This module retrieves sentences relevant to the target SDoH factor from the pre-processed text. First, the Natural Language Toolkit (NLTK) library is used to split the input text into individual sentences~\cite{bird2006nltk}. This pre-processing step is essential, as it allows subsequent modules to focus on smaller, more manageable text segments. Once the input text is split into sentences, a pre-trained LLM (e.g., GPT-3.5-turbo) is used to identify which sentences contain the SDoH factor of interest. 

To achieve this, we crafted a prompt with clear instructions and criteria to evaluate each sentence's relevance to the target SDoH factor. The LLM assigns a relevance label (i.e., Relevant) to sentences that pertain to the factor, and only those deemed relevant are selected for further processing. This step ensures that only the most contextually pertinent sentences are passed on, significantly reducing noise and improving the efficiency of the framework. 

In developing an effective prompt, we tested several variations and evaluated their performance on a curated gold standard development set, using F-1 scores (details of the curated gold standard development set can be found in Section~\ref{sec:data-curation}). The prompt that achieved the highest F-1 score was chosen for use in downstream experiments.

\begin{prompt}
\begin{spverbatim}
[INST]You are a helpful assistant. Read the given context below and find all the sentences that are relevant to `{TARGET_SOCIAL_FACTOR}' based on the provided definition. Format your output (a list of relevant sentences) in a valid JSON payload with one key `Relevant'.[/INST] 

Here is your input: 
[CONTEXT]{INPUT_REPORT}[/CONTEXT]
[{TARGET_SOCIAL_FACTOR}]{FACTOR_DEFINITION}[/{TARGET_SOCIAL_FACTOR}]
\end{spverbatim}
\end{prompt}

\subsubsection{Relevance verification}

Despite extensive prompt engineering, LLMs may occasionally select irrelevant sentences during the retrieval stage, resulting in a flawed set of sentences that appear irrelevant based on human judgment. Inspired by Carupuat et al.~\cite{marasovic-etal-2022-shot}, Weng et al.~\cite{weng2023largelanguagemodelsbetter}, and Gero et al.~\cite{gero2023selfverificationimprovesfewshotclinical}, we introduce an additional verification step using another LLM to re-assess the relevance of the retrieved context. This process involves prompting the LLM with clear instructions to evaluate the sentence's relevance with respect to the target SDoH factor, effectively acting as a human surrogate. We hypothesize that this approach will minimize the inclusion of incorrect or tangential information, ensuring a high-quality set of relevant sentences for downstream processing.

\begin{prompt}
\begin{spverbatim}
[INST]You are a helpful assistant. Verify if the given sentence is relevant to `{TARGET_SOCIAL_FACTOR}' based on the provided definition. Format your output (True or False) in a valid JSON payload with one key `Answer'.[/INST] 

Here is your input: 
[SENTENCE]{TARGET_SENTENCE}[/SENTENCE]
[{TARGET_SOCIAL_FACTOR}]{FACTOR_DEFINITION}[/{TARGET_SOCIAL_FACTOR}]
\end{spverbatim}
\end{prompt}

Despite the impressive performance of LLMs, they often fall short of specialized models in certain tasks. Small models, tailored for a specific task, can be more efficient and faster than LLMs~\cite{bucher2024finetunedsmallllmsstill,lu2023humanwinsllmempirical,edwards2024languagemodelstextclassification}. To evaluate whether comparable relevance verification performance could be achieved with a smaller and more cost-effective LLM, we fine-tuned a FLAN-T5-base model for the binary relevance verification task~\cite{chung2022scalinginstructionfinetunedlanguagemodels}. Given the definition of an SDoH factor and a sentence, the model is tasked to assess the sentence's factual relevance to the factor. 

For fine-tuning, we used a learning rate of 3E-4, a batch size of 8, and conducted 3 epochs. The model was evaluated on the curated test set using F-1 scores, and the fine-tuned model with the highest F-1 score on the test set was selected as the examiner model.

\subsubsection{SDoH factor extraction}

In the final decision-making step, an LLM is used to extract SDoH factors from the verified relevant sentences. At this stage, the focus shifts from identifying relevant context to performing fine-grained information extraction. The verified relevant sentences are passed into the LLM, which identifies and extracts specific SDoH factors occurring within the two weeks preceding the incident, based on the pre-defined definitions outlined in the web coding manual of the National Violent Death Reporting System (Table~\ref{tab:sdoh_factors_definition})\footnote{\url{https://www.cdc.gov/nvdrs/resources/nvdrscodingmanual.pdf}}. 

\begin{prompt}
\begin{spverbatim}
[INST]Read the descriptions and code `True' if any description mentioned 
`{TARGET_SOCIAL_FACTOR}' within the two weeks before the suicide incident. Format your output (True or False) in a valid JSON payload with one key `Happened within two weeks'.[/INST]

Here is your input: 
[CONTEXT]{RELEVANT_DESCRIPTIONS}[/CONTEXT]
[{TARGET_SOCIAL_FACTOR}]{FACTOR_DEFINITION}[/{TARGET_SOCIAL_FACTOR}]
\end{spverbatim}
\end{prompt}

\begin{table}[tbph]
\caption{Suicide-Related SDoH Factor Definitions.}
\label{tab:sdoh_factors_definition}
\small
\rowcolors{2}{}{gray!20!white}

\begin{tabularx}{\textwidth}{>{\raggedright\arraybackslash}p{3cm}X}
\toprule
\textbf{Infrequent Factor} & \textbf{Definition} \\ 
\midrule
Adverse Childhood Experience & Stressful or traumatic events that occurred during childhood and adolescence. \\
Civil Legal Problem & Civil legal (non-criminal) problem(s) appear to have contributed to the death. \\ 
Eviction or Loss of Home & A recent eviction or other loss of the victim’s housing, or the threat of it, appears to have contributed to the death. \\ 
Exposure to Disaster & Exposure to a disaster was perceived as a contributing factor in the incident. \\ 
Financial Problem & Financial problems appear to have contributed to the death. \\ 
Other Addiction & Person has an addiction other than alcohol or other substance abuse, such as gambling, sexual, etc., that appears to have contributed to the death.\\ 
Other Relationship Problem & Problems with a friend or associate (other than an intimate partner or family member) appear to have contributed to the death. \\ 
Other Substance Abuse & Person has a non-alcohol related substance abuse problem. \\ 
Recent Suicide of Friend or Family & Suicide of a family member or friend appears to have contributed to the death. \\ 
School Problem & Problems at or related to school appear to have contributed to the death. \\ 
\bottomrule
\end{tabularx}

\begin{tabularx}{\textwidth}{>{\raggedright\arraybackslash}p{3cm}X}
\textbf{Frequent Factor} & \textbf{Definition} \\ 
\midrule
Alcohol Problem & Person has alcohol dependence or alcohol problems. \\
Criminal Legal Problem & Criminal legal problem(s) appear to have contributed to the death. \\ 
Family Relationship Problem & Victim had relationship problems with a family member (other than an intimate partner) that appear to have contributed to the death.\\
Intimate Partner Problem & Problems with a current or former intimate partner appear to have contributed to the suicide or undetermined death. \\ 
Job Problem & Job problem(s) appear to have contributed to the death.\\ 
Mental Health Problem & Current mental health problem. \\ 
Physical Health Problem & Victim’s physical health problem(s) appear to have contributed to the death. \\ 
Suicide Disclosure & Victims had a history of disclosure of suicidal thoughts or plan. Victim disclosed to another person their thoughts and/or plans to die by suicide. \\ 
\bottomrule
\end{tabularx}
\end{table}

\subsection{Dataset description}
\label{sec:dataset}

The National Violent Death Reporting System (NVDRS) is a population-based active surveillance system that collects data on violent deaths that occur among both residents and non-residents of U.S. states, the District of Columbia, and Puerto Rico~\cite{liu2023nvdrs}. Each incident record includes two death investigation reports: one from a coroner or medical examiner (CME) and another from a law enforcement (LE) reporter. The NVDRS codes over 600 unique data elements for each incident, including SDoH factors that may have affected the victims. These data elements are manually abstracted from the death investigation notes.

For this study, we use the NVDRS 2020 version, which includes 267,804 suicide death incidents, spanning from 2003 to 2019. We focus on 10 infrequent SDoH factors that occurred within the two weeks preceding the suicide incidents (Table~\ref{tab:sdoh_factors_definition}): Adverse Childhood Experience, Civil Legal Problem, Eviction or Loss of Home, Exposure to Disaster, Financial Problem, Other Addiction Problem, Other Relationship Problem, Other Substance Abuse, Recent Suicide of Friends or Family, School Problem. Additionally, to assess whether our framework can match or outperform fine-tuned models on well-represented data, we examine 8 frequent suicide-related SDoH factors occurring within the two weeks leading up to the incident (Alcohol Problem, Criminal Legal Problem, Family Relationship Problem, Intimate Partner Problem, Job Problem, Mental Health Problem, Physical Health Problem, Suicide Disclosure). 

\subsection{Data preparation and curation}
\label{sec:data-curation}
To ensure a fair comparison with the baseline pre-trained model, we adopted the same training and test set split used by Wang et al.~\cite{wang2023sdoh}. To conduct the model evaluation in a cost-effective way, we curated a balanced test set by sampling a balanced number of positive and negative instances for each SDoH factor, and we further sampled a subset for evaluating the DeepSeek-R1 reasoning model (the detailed statistics can be found in Supplementary Table \ref{tab:social_factors_statistics}). This research has been approved by the Institutional Review Board (IRB) at Cornell University. 

To assess the effectiveness of the relevance verification module, we curated a gold-standard dataset, including SDoH factors of interest, human-verified relevant and irrelevant sentences. Two annotators (SW and ML) independently assigned binary relevance labels (`Relevant' and `Not Relevant') to a total of 655 sentences sampled from 160 death investigation notes. These sentences covered 16 suicide-related SDoH factors, except for Adverse Childhood Experience and Suicide Disclosure. Annotators were provided with clear definitions for each SDoH factor to guide their labeling process. In cases of annotation disagreement, the annotators engaged in discussion to reconcile differences and reach consensus. If consensus could not be reached, a third annotator (YP) was consulted for final adjudication. This process ensured the consistent application of the guidelines and the production of high-quality annotations. To evaluate the reliability of the annotation process, we calculated the Inter-Annotator Agreement (IAA) between the annotators. This curated test set was then used to evaluate the effectiveness of both the context retrieval and relevance verification modules.

\subsection{Baselines}
\label{sec:baselines}

We compared our proposed framework against several baseline methods: a fine-tuned BioBERT model~\cite{wang2023sdoh}, a GPT-3.5-turbo End-to-End (End2End) model~\cite{openai2023gpt35}, and a GPT-3.5-turbo Chain-of-Thought (CoT) model~\cite{openai2023gpt35,wei2023chainofthoughtpromptingelicitsreasoning}.

\paragraph{Fine-tuned BioBERT.}

Following Wang et al.~\cite{wang2023sdoh}, we treated SDoH factor extraction as a multi-label text classification task and fine-tuned a BioBERT~\cite{lee2019biobert} model using the NVDRS data. For each victim, we concatenated the CME and LE reports, placing the shorter report first, as input. The model was trained using the Adam optimizer and binary Cross-Entropy loss for parameter optimization. Fine-tuning was conducted with a learning rate of 10E-6, a batch size of 12, and 20 epochs, with early stopping (patience=5) to prevent overfitting. 

The work was performed on an Intel Core i9-9960X 16-core processor, an NVIDIA Quadro RTX 5000 GPU, and 128GB of memory.

\paragraph{GPT-3.5-turbo End2End prompting.}

This baseline employs the GPT-3.5-turbo model~\cite{openai2023gpt35} in an end-to-end fashion, where the system directly maps input narratives to structured SDoH labels without requiring intermediate steps, to extract suicide-related SDoH factors from free text. GPT-3.5-turbo, a large language model developed by OpenAI, is capable of understanding and generating human-like text due to its extensive pre-training on diverse datasets. In this approach, we query GPT-3.5-turbo directly with input text to extract SDoH factors, without additional fine-tuning or few-shot examples. The model processed each input report to generate structured outputs containing the identified SDoH factors. The prompts are carefully crafted with specific instructions to guide the model in focusing on the relevant information within the input text. 

\begin{prompt}
\begin{spverbatim}
[INST]Read the given context below and code `True' if `{TARGET_SOCIAL_FACTOR}' occurred within two weeks before the suicide incident. Format your output (True or False) in a valid JSON payload with one key `Happened within two weeks'.[/INST]

Here is your input: 
[CONTEXT]{INPUT_REPORT}[/CONTEXT]
[{TARGET_SOCIAL_FACTOR}]{FACTOR_DEFINITION}[/{TARGET_SOCIAL_FACTOR}]
\end{spverbatim}
\end{prompt}

\paragraph{GPT-3.5-turbo CoT prompting.}

This baseline utilizes the GPT-3.5-turbo model with a chain-of-thought prompting approach, where LLMs are encouraged to generate intermediate reasoning steps before final prediction. The CoT methodology~\cite{wei2023chainofthoughtpromptingelicitsreasoning} enables the model to generate intermediate reasoning steps, enhancing its ability to handle complex extraction tasks that require multi-step reasoning. The model was instructed to first identify the relevant context related to the target SDoH factor in the input text, and then assess the potential SDoH factor indicators before generating the final answer. The key difference between the CoT approach and the End2End approach lies in the reasoning process. While the End2End method directly produces the final results, the CoT approach explicitly breaks down the reasoning into a chain of thought.

\begin{prompt}
\begin{spverbatim}
[INST]Read the given context below and code `True' if `{TARGET_SOCIAL_FACTOR}' occurred within two weeks before the suicide incident. Answer the following questions one by one: 
1. Is the {TARGET_SOCIAL_FACTOR} mentioned in the context? If yes, please find the relevant sentences. Otherwise, answer 'False'.
2. If the {TARGET_SOCIAL_FACTOR} is mentioned, answer `True' if it happened within two weeks before the suicide incident, answer `False' otherwise.
Format your output in a valid JSON payload with keys `Mentioned or Not', and `Within Two Weeks or Not'[/INST].

Here is your input: 
[CONTEXT]{INPUT_REPORT}[/CONTEXT]
[{TARGET_SOCIAL_FACTOR}]{FACTOR_DEFINITION}[/{TARGET_SOCIAL_FACTOR}]
\end{spverbatim}
\end{prompt}

\subsection{Comparison with the reasoning model}
\label{sec:deepseek}

We also compared extraction performance against the DeepSeek-R1 reasoning model using an end-to-end approach~\cite{deepseekai2025deepseekr1incentivizingreasoningcapability}. Reasoning models, such as DeepSeek-R1, generate extended internal chains of thought, thinking through the problem before providing a final response. The DeepSeek-R1 model was trained via large-scale reinforcement learning without supervised fine-tuning as a preliminary step and showed a remarkable reasoning performance. For this comparison, DeepSeek-R1 was prompted with the input text to extract suicide-related SDoH factors, without additional fine-tuning or provision of few-shot examples. Given the high computational demand for reasoning models, we conducted these comparisons using a subset of the test set, as detailed in Section~\ref{sec:data-curation}.

\begin{prompt}
\begin{spverbatim}
[INST]You will be given a report narrative, and the definition of a social determinants of health factor. Please read the given report, and answer if the factor of interest contributed to the suicide incident within two weeks preceding the suicide incident. Output your reasoning process and your final answer to whether the factor of interest contributed to the suicide incident within two weeks preceding the suicide incident (True or False)". [/INST]

Here is your input: 
[CONTEXT]{INPUT_REPORT}[/CONTEXT]
[{TARGET_SOCIAL_FACTOR}]{FACTOR_DEFINITION}[/{TARGET_SOCIAL_FACTOR}]
\end{spverbatim}
\end{prompt}

\subsection{Evaluating the explainability of the multi-stage design}
\label{sec:method_explainability_evaluation}

In this work, we define explainability as the model's ability to provide explicit, human-interpretable rationales that justify its predictions. More specifically, explainability is characterized by the quality and correctness of the intermediate outputs produced by our framework, namely, the sentences identified by the model as relevant to the SDoH factor being extracted. These intermediate outputs not only reflect the model's reasoning process but also offer transparent evidence supporting its final decision.

To evaluate the explainability of our framework, we focus on two intermediate stages: context retrieval and relevance verification. In the context retrieval stage, the model selects candidate sentences from the incident narrative, while the relevance verification module further filters these candidates to retain only those most pertinent to the queried SDoH factor. We evaluated both stages using our curated gold-standard test set, which includes not only ground-truth SDoH factor labels but also human-verified relevant sentences for each instance. We measured the explainability accuracy by computing the proportion of correctly identified relevant sentences that matched those annotated by human experts.

\subsection{Details of the pilot user study}
\label{sec:explainability_evaluation}

In this work, we conducted a pilot user study to evaluate the practical value of our system for annotating suicide-related SDoH factors. The goal was to assess whether providing human annotators with the framework's intermediate output, highlighting the relevant context, could lead to faster and more accurate annotations. 

To prepare the data, we selected 2 frequently occurring suicide-related SDoH factors (Alcohol Problem and Mental Health Problem) and 2 infrequent factors (Adverse Childhood Experience and Exposure to Disaster). For each factor, we randomly sampled 7 incidents. Participants ($n$=6) were tasked with annotating these 4 factors across the selected incidents twice: once without the relevant context highlighted by our framework (Control arm) and once with the relevant context as AI assistance (Intervention arm). 

We assessed human perceptions by monitoring annotation accuracy and time. These metrics were captured through two anonymous annotation sessions: the first conducted at baseline (Control), and the second conducted one day later (Intervention). This design allowed us to evaluate changes in performance and perceptions over a short time period, while minimizing the impact of task familiarity, providing insights into the effectiveness of our framework.

Following the sessions, participants were asked to complete a questionnaire about their annotation experience and provide feedback for each arm~\cite{sandra2006nasatlx,Bangor2008usability,hoffman2023ai,Jian2000trust}. The specific questions can be found in Figure~\ref{fig:questionnaire}. For Likert scales, we computed the quantitative distribution of responses from the participants for each question.

{
\scriptsize
\setlength{\parindent}{0pt}
\setlength{\parskip}{1em}
\noindent\makebox[\textwidth]{\rule{\textwidth}{0.4pt}}

Q1: On a scale of 1 to 5, how much mental and perceptual activity was required (e.g., thinking, deciding, calculating, remembering, looking, searching, etc.)? 
\begin{enumerate}[nolistsep]
    \item Very little effort
    \item Little effort
    \item Moderate Effort
    \item Considerable effort
    \item Extreme effort
\end{enumerate}

Q2: On a scale of 1 to 5, was the task easy or demanding, simple or complex? 
\begin{enumerate}[nolistsep]
    \item Very easy
    \item Easy
    \item Neutral
    \item Demanding
    \item Very demanding
\end{enumerate}

Q3: On a scale of 1 to 5, how hard did you have to work (mentally and physically) to accomplish this task? 
\begin{enumerate}[nolistsep]
    \item Very little effort
    \item Little effort
    \item Moderate Effort
    \item Considerable effort
    \item Extreme effort
\end{enumerate}

Q4: On a scale of 1 to 5, how successful do you think you were in accomplishing the task?
\begin{enumerate}[nolistsep]
    \item Very unsuccessful
    \item Unsuccessful
    \item Neutral
    \item Successful
    \item Very successful
\end{enumerate}

Q5: On a scale of 1 to 5, how satisfied were you with your performance in accomplishing these goals?
\begin{enumerate}[nolistsep]
    \item Very dissatisfied
    \item Dissatisfied
    \item Neutral
    \item Satisfied
    \item Very satisfied
\end{enumerate}

Q6: On a scale of 1 to 5, how discouraged, stressed versus gratified, relaxed did you feel during the task?
\begin{enumerate}[nolistsep]
    \item Very discouraged
    \item Discouraged
    \item Neutral
    \item Gratified
    \item Very gratified
\end{enumerate}

Q7: This system is reliable.
\begin{enumerate}[nolistsep]
    \item No
    \item Yes
\end{enumerate}

Q8: I feel safe that when I rely on the context provided by the model, I will make the right annotation decisions.
\begin{enumerate}[nolistsep]
    \item No
    \item Yes
\end{enumerate}

Q9: I like using the system for aiding my decision making.
\begin{enumerate}[nolistsep]
    \item No
    \item Yes
\end{enumerate}

Q10: On a scale of 1 to 5, how helpful do you think this type of context is for you to make a QUICK annotation decision?
\begin{enumerate}[nolistsep]
    \item Very unhelpful
    \item Unhelpful
    \item Neutral
    \item Helpful
    \item Very helpful
\end{enumerate}

Q11: On a scale of 1 to 5, how helpful do you think this type of context is for you to make an ACCURATE annotation decision?
\begin{enumerate}[nolistsep]
    \item Very unhelpful
    \item Unhelpful
    \item Neutral
    \item Helpful
    \item Very helpful
\end{enumerate}

Q12: On a scale of 1 to 5, how CONFIDENT are you with your annotations using this type of context?
\begin{enumerate}[nolistsep]
    \item Very unconfident
    \item Unconfident
    \item Neutral
    \item Confident
    \item Very confident
\end{enumerate}
\noindent\makebox[\textwidth]{\rule{\textwidth}{0.4pt}}
\captionsetup{type=figure}
\captionof{figure}{Questionnaire.}
\label{fig:questionnaire}
}

\subsection{Statistics and Reproducibility}
All statistical analyses in this study were performed to evaluate the performance of the proposed multi-stage LLM framework and baseline models for suicide-related SDoH factor extraction. Model performance was assessed using standard metrics, including precision, recall, and F-1 score, computed on balanced test sets comprising both positive and negative instances for each SDoH factor. For evaluation of the context retrieval and relevance verification modules, accuracy was calculated against a curated gold-standard set of 655 sentences, with inter-annotator agreement assessed using Cohen's Kappa statistic (IAA: 91.8\%). The test set sampling, model evaluation, and annotation procedures were predefined and uniformly applied across all experiments to ensure reproducibility. Detailed statistics for dataset splits, sample sizes can be found in Supplementary Table~\ref{tab:social_factors_statistics}.

\section{Results}\label{section:results}

\subsection{Overview of the framework and data sources}

Our framework consists of three steps: context retrieval, relevance verification, and SDoH factor extraction (Figure \ref{fig:overview} and detailed in Section~\ref{sec:framework}). First, the context retrieval stage extracts sentences relevant to the target SDoH factor from the input text. Next, a trained relevance verification module confirms the retrieved context's relevance. Finally, the SDoH factor extraction module determines whether the suicide-related SDoH factor occurred within the two weeks preceding the suicide incident.

\begin{figure}
\includegraphics[width=.8\linewidth]{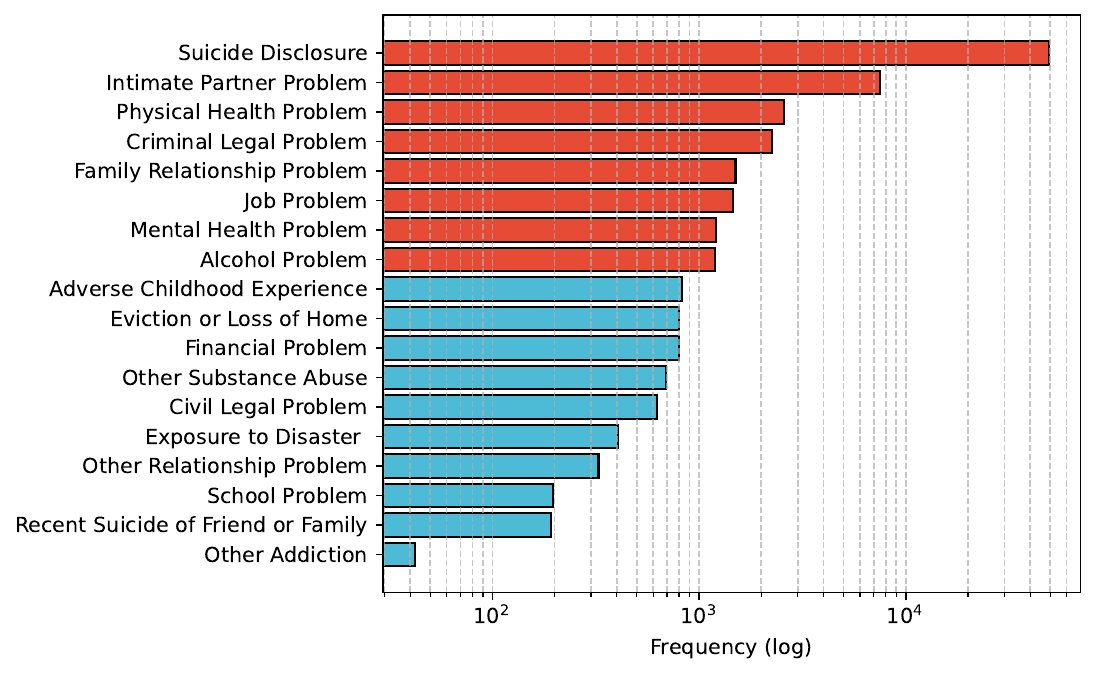}
\caption{Frequency distribution of suicide-related SDoH factors, with the frequent factors shown in red and infrequent factors in blue.}
\label{fig:frequency}
\end{figure}

We evaluated our framework using the National Violent Death Reporting System (NVDRS) dataset~\cite{liu2023nvdrs}, which contains 267,804 reported suicide death incidents across U.S. states, Puerto Rico, and the District of Columbia from 2003 to 2019 (Table~\ref{fig:data} and detailed in Section~\ref{sec:dataset}). The NVDRS provides free-text death investigation notes describing SDoH factors contributing to suicide incidents. While precipitating SDoH factors provide insights into systemic and cumulative risks, immediate pre-incident factors offer a unique lens into the specific triggers and pressures that heighten vulnerability. Although there are over 600 unique SDoH factors documented in NVDRS~\cite{liu2023nvdrs}, our research focuses on 10 infrequent and 8 frequent suicide-related SDoH factors occurring within the two weeks before suicide (Figure \ref{fig:frequency}).

\begin{table}
\caption{Overview of the NVDRS Data.}
\label{fig:data}
\small
\centering
\begin{tabular}{lr}
\toprule
Characteristics & NVDRS \\
\midrule
Total number of subjects & 267,804 \\
Age, mean (SD), year & 46.3 (18.2) \\
Race, \% &  \\
\hspace{1em}American   Indian/Alaska Native & 1.3 \\
\hspace{1em}Asian/Pacific   Islander & 2.3 \\
\hspace{1em}Black or   African American & 6.5 \\
\hspace{1em}Other/Unspecified & 0.9 \\
\hspace{1em}Two or more   races & 1.2 \\
\hspace{1em}Unknown & 0.2 \\
\hspace{1em}White & 87.6 \\
Sex, \% &  \\
\hspace{1em}Female & 22.1 \\
\hspace{1em}Male & 77.9 \\
\hspace{1em}Unknown & 0.0\\
\bottomrule
\end{tabular}
\end{table}

We compared our framework against several baseline models: (1) a fine-tuned BioBERT model~\cite{wang2023sdoh}, (2) a GPT-3.5-turbo model~\cite{openai2023gpt35} in an End-to-End (End2End) approach where the system directly maps input narratives to SDoH labels without intermediate reasoning steps, and (3) a GPT-3.5-turbo model using Chain-of-Thought (CoT) prompting~\cite{openai2023gpt35,wei2023chainofthoughtpromptingelicitsreasoning}, which encourages the LLM to generate explicit intermediate reasoning steps before final prediction (see Section~\ref{sec:baselines} for details). All models were evaluated using a balanced test set for SDoH factor extraction. Additionally, we evaluated our framework against a reasoning model, DeepSeek-R1~\cite{deepseekai2025deepseekr1incentivizingreasoningcapability} using a subset of the test set (Section~\ref{sec:deepseek}). Model performance was assessed using three metrics: precision, recall (sensitivity), and F-1 score (the harmonic mean of the precision and recall). 

To further validate our framework, we measured the accuracy of the relevant context retrieved using a newly curated test set of 160 cases. During curating the test set, Cohen's Kappa score was used to assess annotation agreement, with an Inter-Annotator Agreement (IAA) of 91.8\%, indicating near-perfect agreement based on Mary McHugh's stricter thresholds for healthcare research~\cite{mchugh-interrater}. 

\subsection{Comparison with baseline models}

\subsubsection{Extracting infrequent suicide-related SDoH factors}\label{section:evaluation-infrequent-extraction}

\begin{figure}
\centering
\captionsetup[subfigure]{font={bf,small}, skip=1pt, singlelinecheck=false}
\begin{subfigure}[t]{.32\columnwidth}
\caption{F-1 score}
\label{fig:infrequent_f1}
\includegraphics[width=\linewidth]{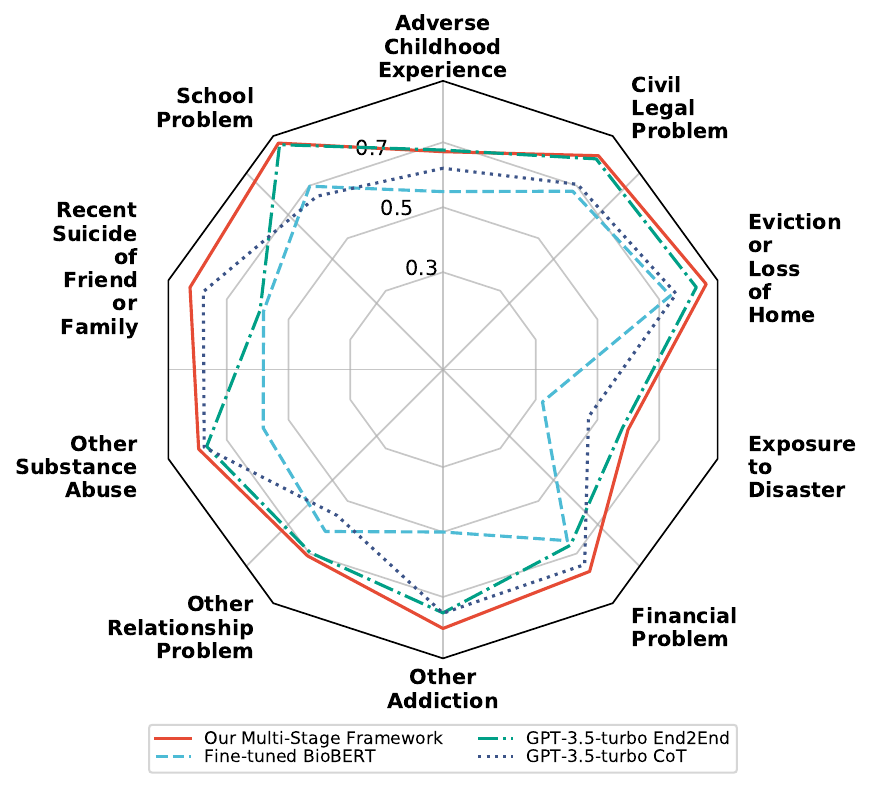}
\end{subfigure}
\hfill
\begin{subfigure}[t]{.32\columnwidth}
\caption{Precision}
\label{fig:infrequent_p}
\includegraphics[width=\linewidth]{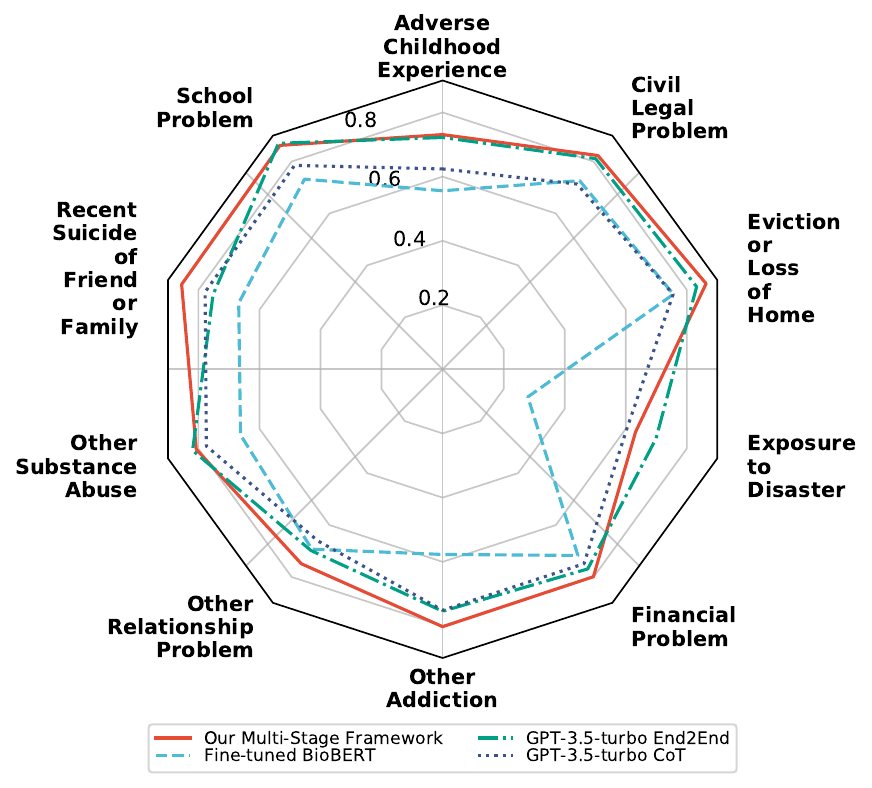}
\end{subfigure}
\hfill
\begin{subfigure}[t]{.32\columnwidth}
\caption{Recall}
\label{fig:infrequent_r}
\includegraphics[width=\linewidth]{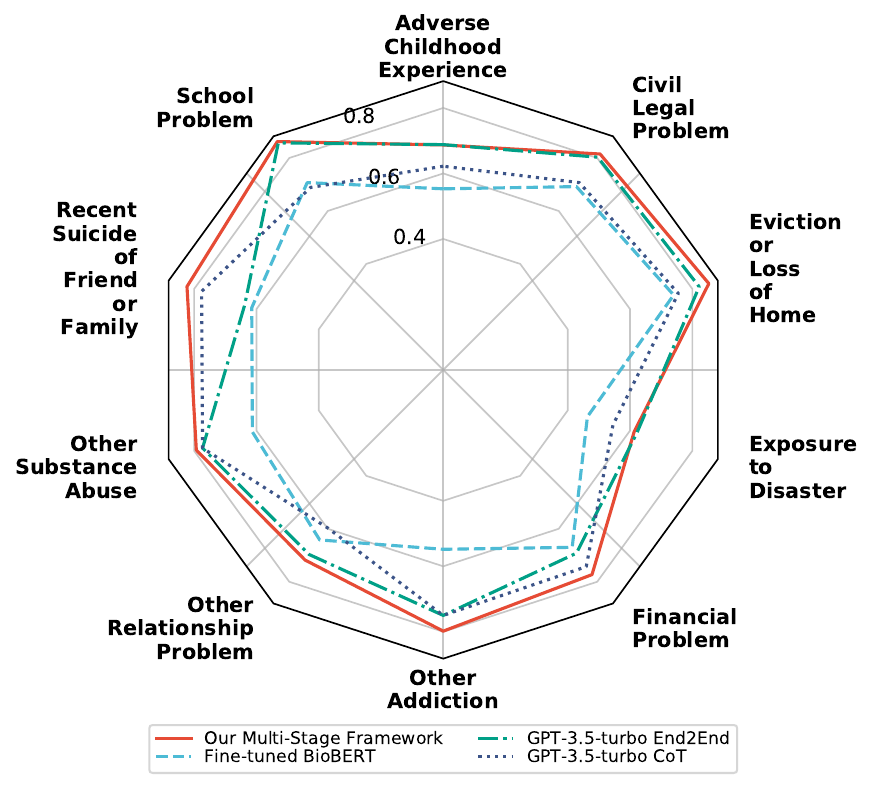}
\end{subfigure}
\caption{Performance comparisons between our proposed multi-stage framework and three baseline models in extracting the 10 infrequent suicide-related SDoH factors from death investigation notes. \textbf{a}, F-1 score comparisons. \textbf{b}, Precision score comparisons. \textbf{c}, Recall score comparisons. CoT - Chain of Thought. Detailed scores can be found in Supplementary Table~\ref{tab:detailed-infrequent-comparisons}.
}
\label{fig:comparison infreq}
\end{figure}

We first evaluated the performance of our framework in extracting 10 infrequent suicide-related SDoH factors (Figure \ref{fig:comparison infreq}). Our framework outperformed all baseline methods in 9 out of 10 infrequent factors. When comparing F-1 scores (Figure \ref{fig:comparison infreq}a), our approach showed an average improvement of 17.7\% over the fine-tuned BioBERT model, 4.8\% over the GPT-3.5-turbo End2End model, and 8.8\% over the GPT-3.5-turbo CoT model.

In terms of precision (Figure \ref{fig:comparison infreq}b), our framework achieved an overall increase of 15.6\% compared to the fine-tuned BioBERT model, 2.0\% compared to the GPT-3.5-turbo End2End model, and 7.3\% compared to the GPT-3.5-turbo CoT model. However, the GPT-3.5-turbo End2End model outperformed our approach in extracting three specific factors: Exposure to Disaster, Other Substance Abuse, and School Problem.

A similar pattern was observed for recall (Figure \ref{fig:comparison infreq}c). Our framework achieved an average recall improvement of 14.9\% over the fine-tuned BioBERT model, 3.9\% over the GPT-3.5-turbo End2End model, and 7.8\% over the GPT-3.5-turbo CoT model. However, the GPT-3.5-turbo End2End model demonstrated superior recall in extracting Adverse Childhood Experience and Exposure to Disaster from the death investigation notes. 

\subsubsection{Extracting frequent suicide-related SDoH factors}\label{section:evaluation-freq-extraction}

\begin{figure}
\centering
\captionsetup[subfigure]{font={bf,small}, skip=1pt, singlelinecheck=false}
\begin{subfigure}[t]{.32\columnwidth}
\caption{F-1 score}
\label{fig:frequent_f1}
\includegraphics[width=\linewidth]{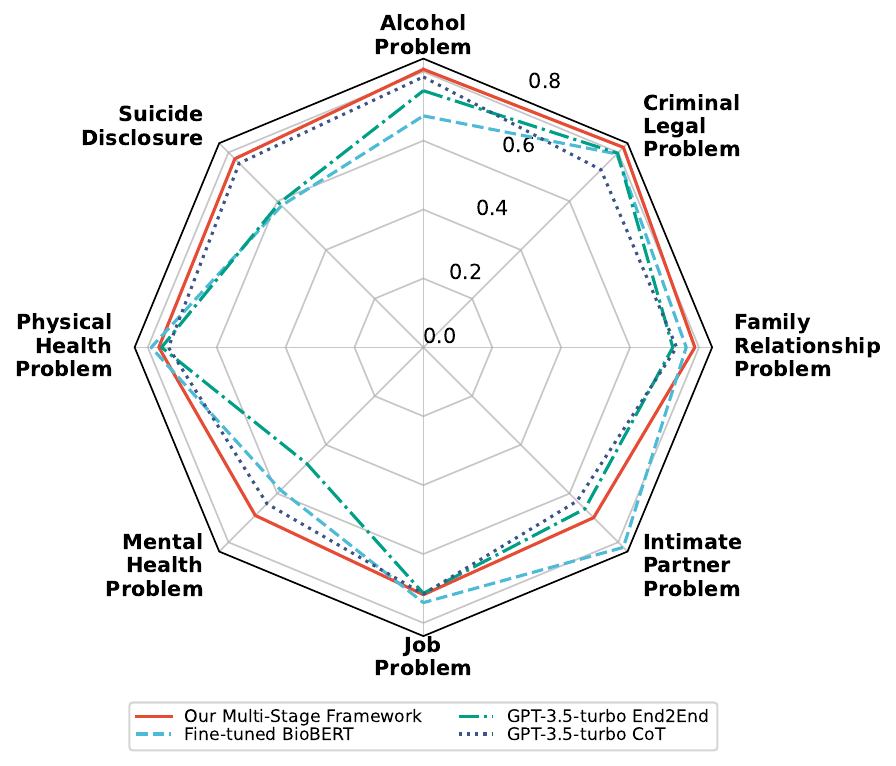}
\end{subfigure}
\hfill
\begin{subfigure}[t]{.32\columnwidth}
\caption{Precision}
\label{fig:frequent_p}
\includegraphics[width=\linewidth]{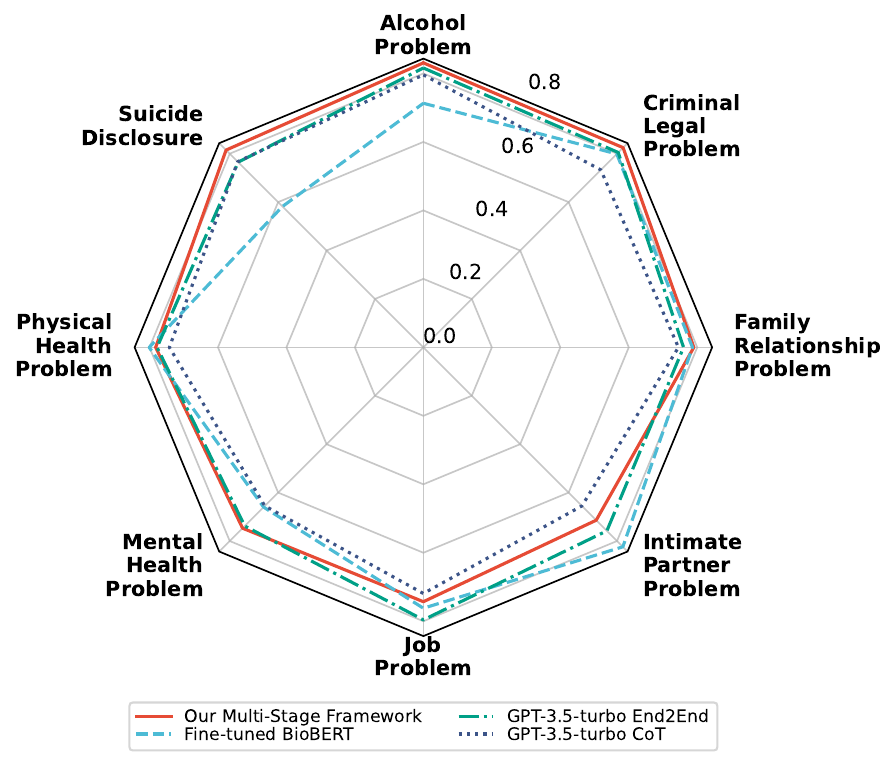}
\end{subfigure}
\hfill
\begin{subfigure}[t]{.32\columnwidth}
\caption{Recall}
\label{fig:frequent_r}
\includegraphics[width=\linewidth]{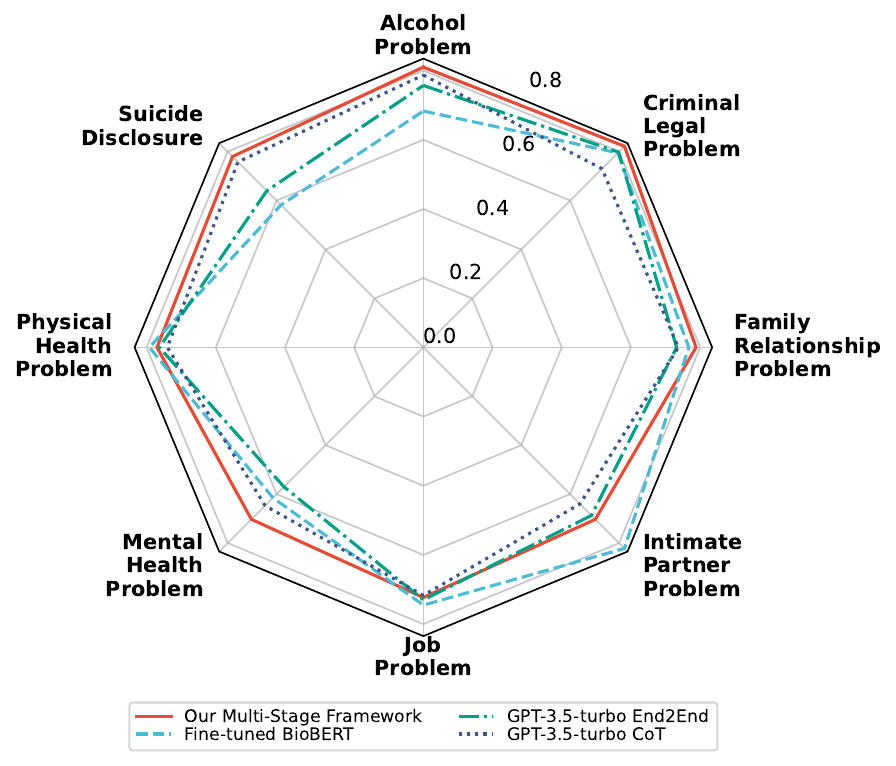}
\end{subfigure}
\caption{Performance comparisons between our proposed multi-stage framework and three baseline models in extracting the 8 frequent suicide-related SDoH factors from death investigation notes. \textbf{a}, F-1 score comparisons. \textbf{b}, Precision score comparisons. \textbf{c}, Recall score comparisons. CoT - Chain of Thought. Detailed scores can be found in Supplementary Table~\ref{tab:detailed-frequent-comparisons}.}
\label{fig:comparison freq}
\end{figure}

We then compared our multi-stage framework with baselines for extracting 8 frequent suicide-related SDoH factors from death investigation notes (Figure \ref{fig:comparison freq}). Our framework consistently improved F-1 scores for 5 out of 8 factors (Figure \ref{fig:comparison freq}a). Notably, despite operating in a zero-shot setting without fine-tuning, our framework achieved an average F-1 score improvement of 4.0\% over the fine-tuned BioBERT model. It also outperformed the zero-shot GPT-3.5-turbo End2End and CoT models by 7.4\% and 4.1\%, respectively. 

Similar trends were observed in precision and recall scores (Figure \ref{fig:comparison freq}b and Figure \ref{fig:comparison freq}c). Our framework improved precision by 4.0\% over the fine-tuned BioBERT model, 0.4\% over the GPT-3.5-turbo End2End model, and 5.4\% over the GPT-3.5-turbo CoT model. Additionally, it enhanced recall by 3.7\% compared to BioBERT, 5.3\% compared to GPT-3.5-turbo End2End, and 4.3\% compared to GPT-3.5-turbo CoT.

\subsection{Comparison with the reasoning model}

\begin{figure}
\centering
\captionsetup[subfigure]{font={bf,small}, skip=1pt, singlelinecheck=false}
\begin{subfigure}[t]{.32\columnwidth}
\caption{F-1 score}
\label{fig:deepseek_f1}
\includegraphics[width=\linewidth]{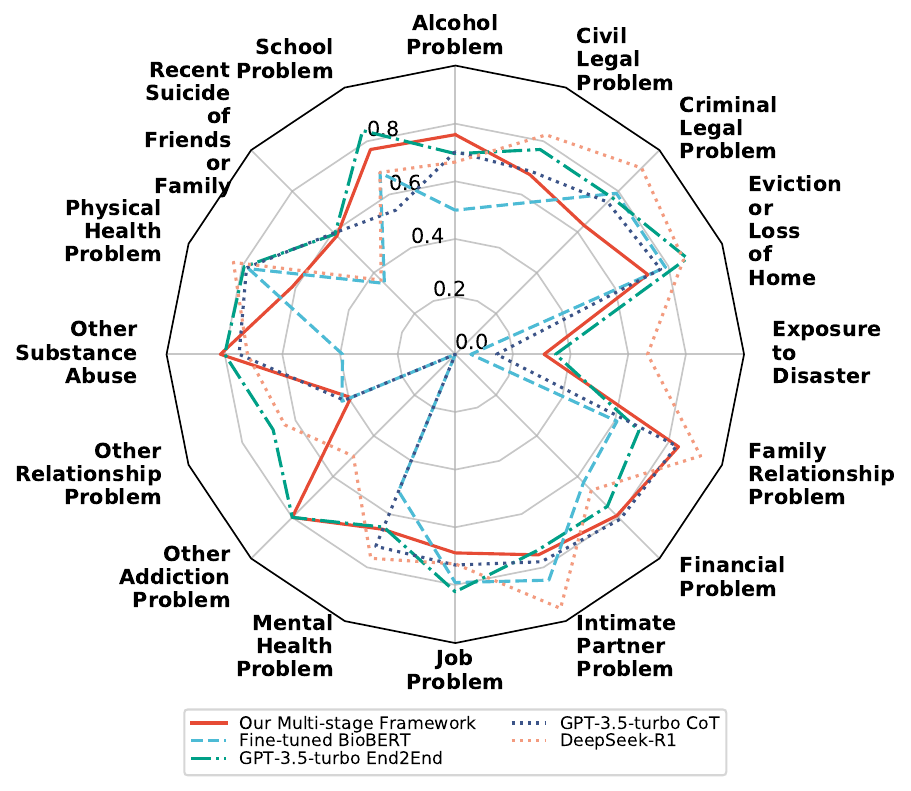}
\end{subfigure}
\hfill
\begin{subfigure}[t]{.32\columnwidth}
\caption{Precision}
\label{fig:deepseek_p}
\includegraphics[width=\linewidth]{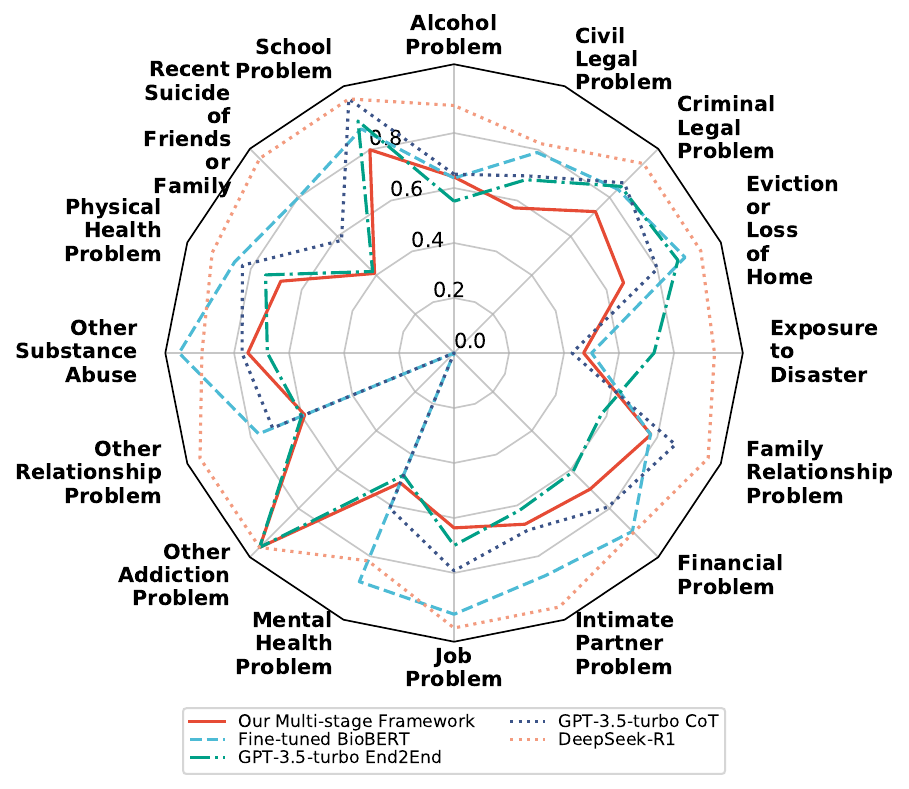}
\end{subfigure}
\hfill
\begin{subfigure}[t]{.32\columnwidth}
\caption{Recall}
\label{fig:deepseek_r}
\includegraphics[width=\linewidth]{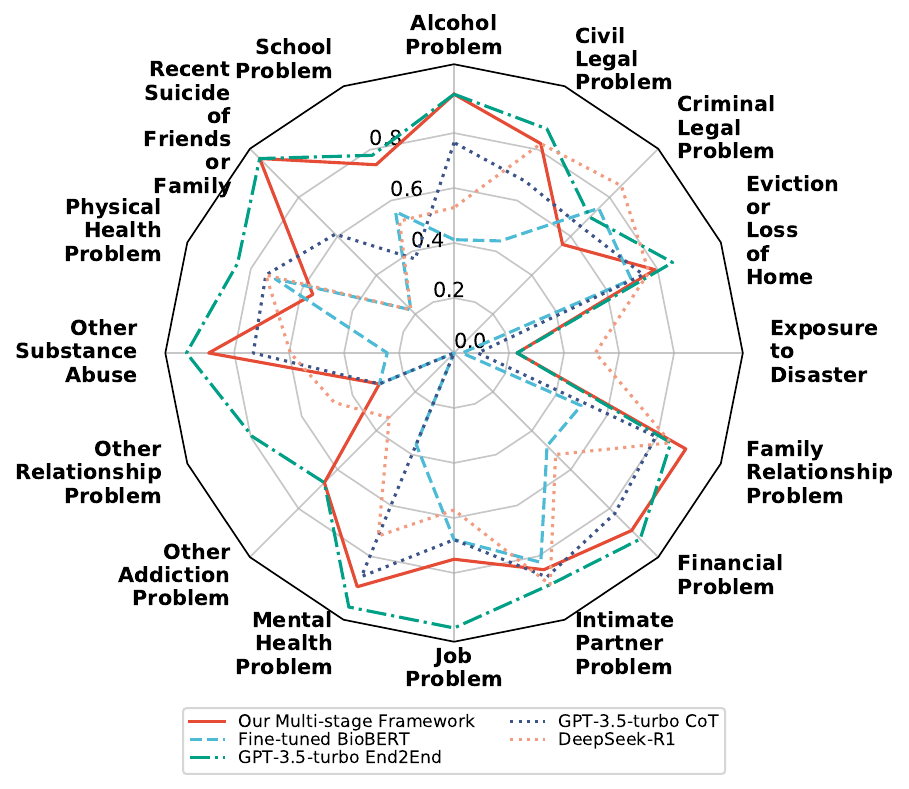}
\end{subfigure}
\caption{Performance comparisons between our proposed multi-stage framework, baseline models, and the DeepSeek-R1 reasoning model in extracting suicide-related SDoH factors from death investigation notes. \textbf{a}, F-1 score comparisons. \textbf{b}, Precision score comparisons. \textbf{c}, Recall score comparisons. CoT - Chain of Thought. Detailed scores can be found in Supplementary Table~\ref{tab:detailed-comparisons-deepseek}.}
\label{fig:comparison deepseek}
\end{figure}

We compared the performances of our proposed framework with the DeepSeek-R1 reasoning model on a subset of the SDoH factor extraction test set (Figure \ref{fig:comparison deepseek}) due to the higher computational demands of these models. As shown in Figure \ref{fig:comparison deepseek}a, DeepSeek-R1 outperformed other models in 7 out of 16 evaluated SDoH factors. Notably, it achieved an F-1 score of 66.7\% on the infrequent Exposure to Disaster factor, which is 35.9\% higher than our framework, 31.9\% higher than the GPT-3.5-turbo End2End model, and 61.3\% higher than the fine-tuned BioBERT model. In terms of precision scores (Figure \ref{fig:comparison deepseek}b), DeepSeek-R1 achieved the highest precision in 14 out of 16 factors. However, Figure \ref{fig:comparison deepseek}c shows that DeepSeek-R1 achieved the highest recall score in only 3 out of 16 factors, while our proposed framework outperformed it on recall for 11 factors.

\subsection{Evaluating the explainability of the multi-stage design}\label{section:result-effectiveness-evaluation}

We evaluated the context retrieval and relevance verification stages of our framework using the curated test set, as these intermediate outputs directly inform the model's final decisions. Crucially, both stages produce interpretable outputs, which are the selected contextual evidence that serve as explicit intermediate explanations for each prediction. This multi-layered decision-making process enhances the framework's explainability by transparently revealing how the model reasons through retrieved evidence before arriving at a final decision.

\begin{figure}
    \centering
    \includegraphics[width=.5\linewidth]{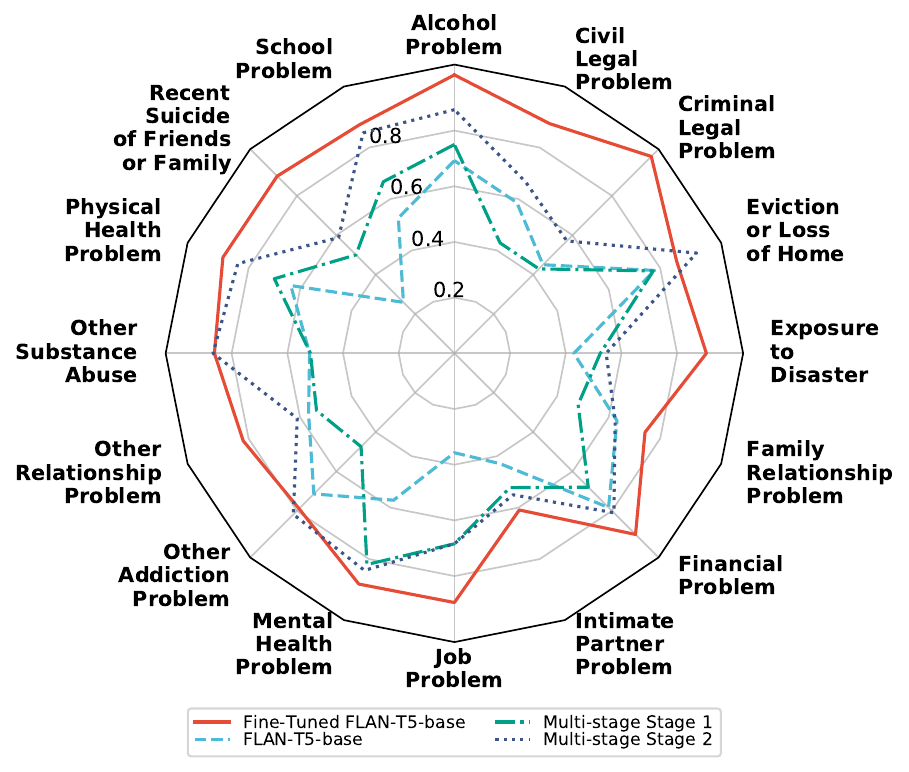}
    \caption{Accuracy score comparisons between the context retrieval in stage 1 and the relevance verification in stage 2, to examine the effectiveness of our multi-stage design in the proposed framework. Detailed scores can be found in Supplementary Table~\ref{tab:detailed-accuracy-comparisons}.}
    \label{fig:accuracy}
\end{figure}

To assess the effectiveness of this pipeline, we measured the accuracy of the retrieved context at each stage. As shown in Figure \ref{fig:accuracy}, our relevance verification process improved overall relevant context retrieval accuracy from 59.3\% to 73.1\%, representing a 13.8\% improvement over the single-stage context retrieval approach. Additionally, we assessed the FLAN-T5-base model's performance in the relevance verification task. Without fine-tuning, it achieved an average accuracy of 55.8\%. After fine-tuning, the accuracy increased by 31\%, reaching 86.6\%.

\subsection{Pilot user study}

We conducted a pilot user study to assess the practical value of our system for annotating SDoH factors. The primary objective was to validate that experts using our proposed method could annotate suicide-related SDoH factors more quickly and accurately than through manual efforts alone. To test this claim, we employed a two-arm design: one involving experts working independently (Control) and the other combining expert efforts with AI assistance (Intervention). We recruited 6 human experts to annotate 4 suicide-related SDoH factors (Adverse Childhood Experience, Alcohol Problem, Exposure to Disaster, Mental Health Problem) from 28 reports. Initially, the experts annotated the SDoH factors independently without AI assistance (Control). After at least 24 hours, they annotated the same set of factors with AI assistance (Intervention) to minimize the effects of task familiarity. After completing the annotations, experts were asked to fill out a survey to provide feedback.

\begin{figure}
\centering
\begin{subfigure}[t]{\columnwidth}
\includegraphics[width=\linewidth]{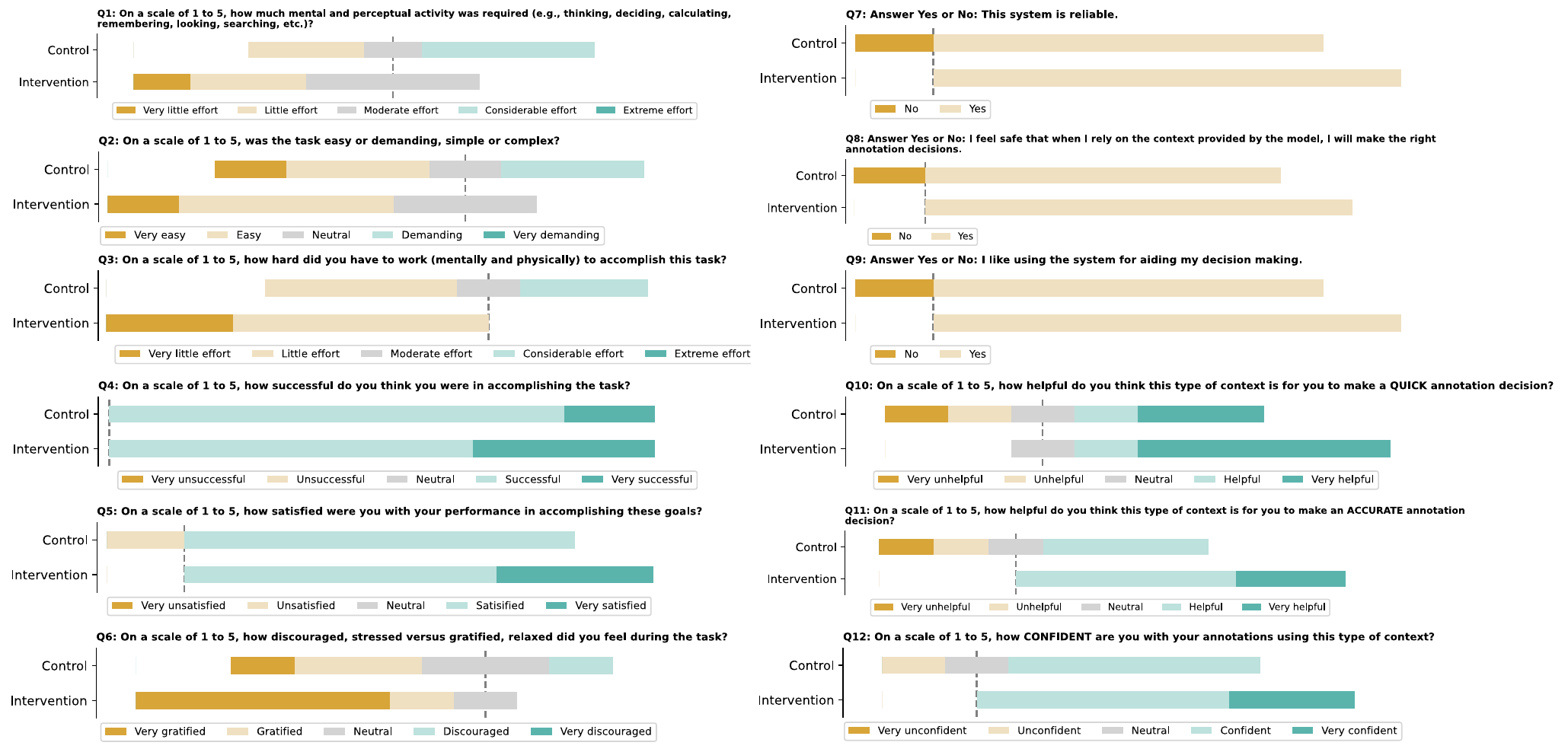}
\caption{User Study Survey Results}
\label{fig:human_study_questionnaire}
\end{subfigure}
\begin{subfigure}[t]{.44\columnwidth}
\centering
\includegraphics[width=\linewidth]{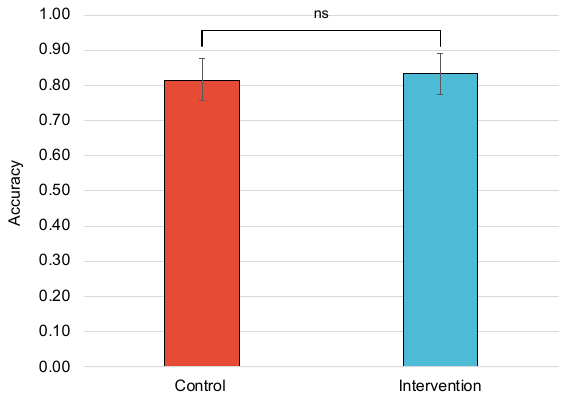}
\caption{Accuracy}
\label{fig:human_study_accuracy}
\end{subfigure}
\begin{subfigure}[t]{.4\columnwidth}
\centering
\includegraphics[width=\linewidth]{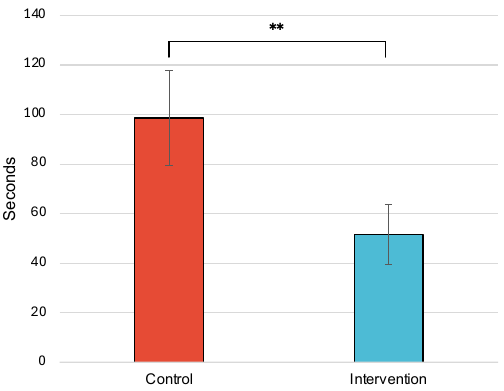}
\caption{Annotation Time}
\label{fig:human_study_annotation_time}
\end{subfigure}
\caption{Pilot user studies for suicide-related SDoH factor annotation. \textbf{a}, user study survey quantitative response distributions. \textbf{b}, the annotation accuracy comparisons between the control arm and the intervention arm. \textbf{c}, the annotation time comparisons between the control arm and the intervention arm.}
\label{fig:pilot}
\end{figure}

Figure \ref{fig:pilot}a illustrates that experts experienced a lower mental and/or physical stress when annotating with AI assistance compared to working independently (Q1-Q4). They also reported higher satisfaction with their annotation performance (Q5). When asked about system reliability and their interest in using the system for decision-making support (Q7-Q9), all participants expressed a preference for the AI-assisted method. Furthermore, participants noted that the AI assistance resulted in quicker, more accurate, and more confident annotations compared to working independently (Q10-Q12).

Figure \ref{fig:pilot}b shows that experts who collaborated with AI assistance achieved an annotation accuracy of 83.33\%, slightly surpassing the 81.55\% accuracy achieved by experts working independently. Similarly, Figure \ref{fig:pilot}c shows that annotating with AI assistance required less time, saving an average of 62.39 seconds per incident compared to independent annotation. 

\section{Discussion}\label{section:discussion}

Our work presents an approach using LLMs to extract suicide-related SDoH factors from unstructured text. The framework breaks down the extraction task into three key steps: context retrieval, relevance verification, and SDoH factor extraction. This multi-stage approach not only outperforms baseline models but also enhances transparency by providing intermediate outputs that shed light on the decision-making process.

Comparisons with baseline methods underscore the effectiveness of our proposed framework. Our approach consistently improved F-1 scores for 9 out of 10 infrequent suicide-related SDoH factors and 5 out of 8 frequent suicide-related SDoH factors occurring within the two weeks prior to the suicide incidents. Specifically, for infrequent factors, the average F-1 score of our framework exceeded the fine-tuned BioBERT model by 17.7\%, the GPT-3.5-turbo End2End model by 4.8\%, and the GPT-3.5-turbo CoT model by 8.8\%. For frequent factors, the average F-1 score of our zero-shot multi-stage framework was 4.0\% higher than the fine-tuned BioBERT model, 7.4\% higher than the GPT-3.5-turbo End2End model, and 4.1\% higher than the GPT-3.5-turbo CoT model. Additionally, we demonstrated that the relevance verification step improved the overall relevant context retrieval accuracy, increasing it from 59.3\% to 73.1\%, marking a 13.8\% improvement over the single-stage context retrieval. Our study highlights the distinct ways in which different modeling techniques address the challenge of long-tailed SDoH distributions. The fine-tuned BioBERT model, though specifically optimized for extracting SDoH factors from text, showed limited generalizability to low-frequency SDoH factors. This limitation reflects the model's reliance on the empirical distribution of the training data and its inherent sensitivity to class imbalance. In contrast, LLM-based approaches, which utilize instruction prompting and benefit from large-scale pretraining, demonstrated better robustness. These models were more effective in extracting low-frequency SDoH factors, even in zero-shot settings.

The differences in performance improvements observed across our model comparisons, particularly the more substantial gains over baseline models compared to reasoning models, highlight important considerations regarding model capabilities and practical deployment. Our framework demonstrated strong improvements over baseline models across multiple SDoH factors, especially in handling infrequent factors. When compared to reasoning models, performance gains were more modest and nuanced: our model achieved higher recall, while the reasoning models exhibited higher precision. These contrasting patterns suggest that the two approaches may be optimized for different application needs, where our framework is better suited for high-coverage extraction tasks. In contrast, reasoning models may be preferable when minimizing false positives is critical. It is also important to note that the evaluation sets used in each comparison differed. A smaller, more controlled set was used for reasoning models due to their higher computational demands. These evaluation constraints likely contribute to the observed differences in performance improvements and should be taken into consideration when interpreting the results.

The pilot user study further validated that human experts using the intermediate explanations provided by our framework reduced annotation time by an average of 62.39 seconds without sacrificing accuracy, maintaining an annotation accuracy of 83.33\%.

However, this study has several limitations. First, as shown in Section~\ref{section:result-effectiveness-evaluation}, fine-tuning a small FLAN-T5-base model improved the average accuracy of our multi-stage relevant context retrieval by 13.7\%. However, unlike the GPT-3.5-turbo, the fine-tuned FLAN-T5 model does not inherently produce reasoning behind its predictions, limiting its explainability. To enable such reasoning capability, human-generated reasoning would need to be incorporated during the fine-tuning process, which was not implemented in this study. 
Second, we evaluated our model using a selection of 10 infrequent and 8 frequent SDoH factors, which represent only a subset of the full range of factors defined in the NVDRS coding manual~\cite{liu2023nvdrs}. By including both high-frequency and low-frequency factors, we aimed to evaluate the model's performance under varying data availability scenarios and assess its potential generalizability to the more imbalanced distributions often encountered in real-world settings. The selection of these 18 factors was guided by both their observed frequency in the NVDRS data and their relevance as identified in existing literature on suicide risk~\cite{dang2023-suicide-ontology, wang2023sdoh}. It is important to note, however, that the NVDRS encompasses over 600 unique SDoH-related factors, many of which may hold clinical or contextual significance. Expanding this framework in future research to incorporate a broader spectrum of SDoH factors would further enhance the comprehensiveness of information extraction and improve support for a wide range of downstream public health applications. 
While the NVDRS provides a rich and detailed source of information on violent deaths, including suicide, it is important to recognize that certain SDoH factors may be more prevalent or more frequently documented among specific subpopulations due to underlying societal and structural influences. Such differences may introduce variability in model performance across demographic groups. Moreover, LLMs are trained on datasets that may not equally represent the full range of linguistic, cultural, or socioeconomic contexts, which could further affect the consistency and fairness of their outputs. These considerations highlight the need for future research to systematically evaluate model fairness across subpopulations and assess performance disparities.

Despite growing interest in explainable NLP, there is currently no universal definition or standard evaluation framework for explainability that applies across diverse tasks and model architectures~\cite{lyu-etal-2024-towards,2024-zhao-explainability}. In practice, explainability is often approximated through proxy measures such as feature importance, attention weights, or the presence of interpretable intermediate outputs. However, these proxies do not consistently align with human reasoning or trust, and their reliability as true explanations remains debated. This challenge is particularly true for LLMs, whose internal decision-making processes are largely opaque and highly contextual. In this work, we address these challenges by introducing a multi-stage framework where intermediate outputs, such as retrieved and verified context, serve as explicit and human-interpretable rationales. While our approach enhances transparency in machine-learning models, we recognize that explainability remains inherently task-specific and subjective. Further efforts are needed to develop standardized, user-centered frameworks for evaluating and comparing explanation quality across different models and domains.

Although our pilot user study demonstrated reduced annotation time when LLM-generated intermediate explanations were provided, the within-subject design may introduce potential confounding effects related to task familiarity. Annotators completed the intervention task one day after the control task. While sessions were separated by 24 hours to minimize carryover effects~\cite{CHARNESS20121}, increased familiarity with the annotation task may still have contributed to the observed efficiency gain. To more rigorously evaluate the causal impact of AI assistance on annotation quality and efficiency, future studies should adopt a between-subjects design (e.g., comparing a group exposed to AI assistance with a separate group completing repeated control tasks). Therefore, the findings of this study should be interpreted as exploratory rather than conclusive.

\section*{Data Availability}
The dataset analyzed in this study, the NVDRS Restricted Access Database (RAD), is accessible upon request to researchers who meet specific eligibility criteria and take steps to ensure confidentiality and data security. Our research was approved by the NVDRS RAD proposal, which gave us the required permissions to access the data and undertake the work described here. This restricted access is in place due to the confidential nature of the NVDRS data, which includes sensitive information that could potentially lead to the unintended disclosure of the identities of victims. To safeguard this data, the CDC protects it by requiring users to fulfill certain eligibility requirements and implement the necessary measures to ensure the security of data, preserve confidentiality, and prevent unauthorized access. Researchers interested in accessing the NVDRS data can apply per the instructions provided at \url{https://www.cdc.gov/nvdrs}. 

\section*{Code Availability}
Our code and implementations will be made publicly available at \url{https://github.com/bionlplab/2025_multistage_llm_sdoh}.


\section*{Acknowledgements}

This study was supported by the AIM-AHEAD Consortium Development Program of NIH under grant number OT2OD032581. The content is solely the responsibility of the authors and does not necessarily represent the official views of the NIH.

\section*{Author Contributions}
S.W. and Y.P. contributed to the conception of the study and study design; S.W., K.D., M.L., H.M., Y.M., Y.W., Z.X., J.Z., Y.X., Y.P. contributed to the acquisition of the data; S.W. and Y.P. contributed to the analysis and interpretation of the data; Y.X., Y.D., X.X., J.G., and Y.P. provided strategic guidance; S.W. and Y.P. contributed to the paper organization and team logistics; S.W., K.D., M.L., H.M., Y.M., Y.W., Z.X., J.Z., Y.X., Y.D., X.X., J.G., and Y.P. contributed to drafting and revising the manuscript.

\section*{Ethics declarations \& Competing Interests}
The authors declare no competing interests.

\newpage
\setlength{\bibsep}{0pt plus 0.3ex}
\bibliographystyle{unsrtnat}
\bibliography{preprint}

\newpage
\appendix
\setcounter{subsection}{0}
\setcounter{table}{0}
\setcounter{figure}{0}
\renewcommand{\thesubsection}{S\arabic{subsection}}
\renewcommand\figurename{Supplementary Figure} 
\renewcommand\tablename{Supplementary Table}

\begin{table}[!hbpt]
\caption{SDoH factor statistics in the training and test sets.}
\label{tab:social_factors_statistics}
\centering
\begin{tabular}{lrrrrrr}
\toprule
& \multicolumn{2}{c}{Training} & \multicolumn{2}{c}{Test} & \multicolumn{2}{c}{\makecell{Testing\\DeepSeek-R1}} \\ 
\cmidrule(rl){2-3}\cmidrule(rl){4-5}\cmidrule(rl){6-7}
Factor & {Positive} & {Negative}  & {Positive} & {Negative}  & {Positive} & {Negative} \\ 
\midrule
Adverse Childhood Experience & 831 & 88,284 & 300 & 300 & - & - \\
Alcohol Problem & 1,197 & 87,918 & 300 & 300 & 17 & 25\\
Civil Legal Problem & 626 & 88,489 & 300 & 300 & 34 & 41\\ 
Criminal Legal Problem & 2,252 & 86,863 & 300 & 300 & 43 & 37\\ 
Eviction or Loss of Home & 800 & 88,315 & 300 & 300 & 43 & 36 \\ 
Exposure to Disaster & 404 & 88,711 & 300 & 300 & 35 & 35\\ 
Family Relationship Problem & 1,500 & 87,615 & 300 & 300 & 34 & 39 \\
Financial Problem & 797 & 88,318 & 300 & 300 & 23 & 21 \\ 
Intimate Partner Problem & 7,544 & 81,571 & 300 & 300 & 34 & 31\\ 
Job Problem & 1,453 & 87,662 & 300 & 300 & 28 & 25 \\ 
Mental Health Problem & 1,210 & 87,905 & 300 & 300 & 25 & 31\\ 
Other Addiction & 42 & 89,073 & 42 & 42 & 3 & 1 \\ 
Other Relationship Problem & 326 & 88,789 & 300 & 300 & 34 & 36 \\ 
Other Substance Abuse & 695 & 88,420 & 300 & 300 & 37 & 30\\ 
Physical Health Problem & 2,589 & 86,526 & 300 & 300 & 27 & 21 \\ 
Recent Suicide of Friend or Family & 192 & 88,923 & 192 & 192 & 18 & 34 \\ 
School Problem & 196 & 88,919 & 196 & 196 & 27 & 19 \\ 
Suicide Disclosure & 49, 140 & 39,975 & 300 & 300 & - & - \\ 
\bottomrule
\end{tabular}

\end{table}

\newpage

\begin{table}[!hbpt]
\caption{Detailed comparisons between our proposed multi-stage framework and baseline models in extracting the 10 infrequent suicide-related SDoH factors from death investigation notes. P - Precision, R - Recall.}
\label{tab:detailed-infrequent-comparisons}
\resizebox{\textwidth}{!}{
\begin{tabular}{lrrrrrrrrrrrr}
\toprule
& \multicolumn{3}{c}{Fine-tuned BioBERT} & \multicolumn{3}{c}{\makecell[c]{GPT-3.5-turbo\\End2End}} & \multicolumn{3}{c}{GPT-3.5-turbo CoT} & \multicolumn{3}{c}{\makecell[c]{Our Multi-Stage\\Framework}} \\
\cmidrule(rl){2-4}\cmidrule(rl){5-7}\cmidrule(rl){8-10}\cmidrule(rl){11-13}
Factor & P & R & F-1 & P & R & F-1 & P & R & F-1 & P & R & F-1 \\
\midrule
Adverse Childhood Experience  & 0.556 & 0.553 & 0.548 & 0.722 & 0.688 & 0.676 & 0.624 & 0.622 & 0.620 & 0.731 & 0.687 & 0.671 \\
Civil Legal Problem & 0.726 & 0.692 & 0.679 & 0.811 & 0.803 & 0.802 & 0.713 & 0.707 & 0.705  & 0.823 & 0.815 & 0.814 \\
Eviction or Loss of Home & 0.756 & 0.740 & 0.736 & 0.832 & 0.822 & 0.820 & 0.755 & 0.755 & 0.755 & 0.863 & 0.853 & 0.852 \\
Exposure to Disaster & 0.279 & 0.463 & 0.322 & 0.699 & 0.622 & 0.581 & 0.603 & 0.545 & 0.471 & 0.632 & 0.613 & 0.599 \\
Financial Problem & 0.717  & 0.670 & 0.651 & 0.769 & 0.692 & 0.668      & 0.749 & 0.743 & 0.742 & 0.799 & 0.773 & 0.768 \\
Other Addiction & 0.577 & 0.548 & 0.500 & 0.754 & 0.750 & 0.749 & 0.751 & 0.750 & 0.750 & 0.802 & 0.798 & 0.797 \\
Other Relationship Problem & 0.693 & 0.642 & 0.616 & 0.698  & 0.697 & 0.696 & 0.661 & 0.598 & 0.555 & 0.749 & 0.718 & 0.709 \\
Other Substance Abuse & 0.662 & 0.613 & 0.582 & 0.819 & 0.773 & 0.765 & 0.774 & 0.773 & 0.773 & 0.807 & 0.793 & 0.791 \\
Recent Suicide of Friend or Family & 0.668 & 0.615 & 0.581 & 0.752 & 0.638 & 0.592 & 0.778 & 0.776 & 0.776 & 0.855 & 0.823 & 0.819 \\
School Problem & 0.732 & 0.707 & 0.698 & 0.870 & 0.857 & 0.856  & 0.785  & 0.689 & 0.660 & 0.862  & 0.862 & 0.862 \\
\bottomrule
\end{tabular}
}
\end{table}

\newpage

\begin{table}[!hbpt]
\caption{Detailed comparisons between our proposed multi-stage framework and baseline models in extracting the 8 frequent suicide-related SDoH factors from death investigation notes. P - Precision, R - Recall.}
\label{tab:detailed-frequent-comparisons}
\resizebox{\textwidth}{!}{
\begin{tabular}{lrrrrrrrrrrrr}
\toprule
& \multicolumn{3}{c}{Fine-tuned BioBERT} & \multicolumn{3}{c}{\makecell[c]{GPT-3.5-turbo\\End2End}} & \multicolumn{3}{c}{GPT-3.5-turbo CoT} & \multicolumn{3}{c}{\makecell[c]{Our Multi-Stage\\Framework}} \\
\cmidrule(rl){2-4}\cmidrule(rl){5-7}\cmidrule(rl){8-10}\cmidrule(rl){11-13}
Factor & P & R & F-1 & P & R & F-1 & P & R & F-1 & P & R & F-1 \\
\midrule
Alcohol Problem	& 0.713	& 0.683	& 0.672	& 0.816	& 0.757	& 0.745	& 0.796	& 0.787	& 0.785	& 0.831	& 0.810	& 0.807 \\
Criminal Legal Problem	& 0.799	& 0.795	& 0.794	& 0.805	& 0.798	& 0.797	& 0.733	& 0.730	& 0.729	& 0.825	& 0.822	& 0.821 \\
Family Relationship Problem	& 0.786	& 0.767	& 0.763	& 0.760	& 0.732	& 0.724	& 0.745	& 0.737	& 0.735	& 0.790	& 0.788	& 0.788 \\
Intimate Partner Problem & 0.824 & 0.822 & 0.821 & 0.758 & 0.687 & 0.663 & 0.655 & 0.640 & 0.631 & 0.714 & 0.703 & 0.699 \\
Job Problem	& 0.762	& 0.745	& 0.741	& 0.796	& 0.730	& 0.714	& 0.719	& 0.717 & 0.716 & 0.743 & 0.723 & 0.718 \\
Mental Health Problem & 0.660 & 0.615 & 0.586 & 0.738 & 0.570 & 0.478 & 0.655 & 0.647 & 0.642 & 0.747 & 0.703 & 0.690 \\
Physical Health Problem	& 0.801	& 0.792	& 0.790	& 0.778	& 0.763	& 0.760	& 0.742	& 0.740	& 0.740	& 0.782	& 0.770	& 0.768 \\ 
Suicide Disclosure	& 0.582	& 0.582	& 0.582	& 0.765	& 0.640	& 0.592	& 0.767	& 0.758	& 0.756	& 0.815	& 0.780	& 0.774 \\ 
\bottomrule
\end{tabular}
}
\end{table}

\newpage

\begin{table}[!hbpt]
\caption{Detailed performance comparisons between our proposed multi-stage framework, baseline models, and the DeepSeek-R1 reasoning model in extracting the suicide-related SDoH factors from death investigation notes. P - Precision, R - Recall.}
\label{tab:detailed-comparisons-deepseek}
\resizebox{\textwidth}{!}{
\begin{tabular}{lrrrrrrrrrrrrrrr}
\toprule
& \multicolumn{3}{c}{Fine-tuned BioBERT} & \multicolumn{3}{c}{\makecell[c]{GPT-3.5-turbo\\End2End}} & \multicolumn{3}{c}{GPT-3.5-turbo CoT} & \multicolumn{3}{c}{\makecell[c]{Our Multi-Stage\\Framework}} & \multicolumn{3}{c}{\makecell[c]{DeepSeek-R1}}\\
\cmidrule(rl){2-4}\cmidrule(rl){5-7}\cmidrule(rl){8-10}\cmidrule(rl){11-13}\cmidrule(rl){14-16}
Factor & P & R & F-1 & P & R & F-1 & P & R & F-1 & P & R & F-1 & P & R & F-1\\
\midrule
Adverse Childhood Experience  & - & - & - & - & - & - & - & - & - & - & - & - & - & - & - \\
Alcohol Problem	& 0.636 & 0.412 & 0.500 & 0.552 & 0.941 & 0.696 & 0.650 & 0.765 & 0.703 & 0.640 & 0.941 & 0.762 & 0.900 & 0.529 & 0.667 \\
Civil Legal Problem  & 0.789 & 0.441 & 0.566 & 0.682 & 0.882 & 0.769 & 0.697 & 0.676 & 0.687 & 0.571 & 0.824 & 0.675 & 0.824 & 0.824 & 0.824 \\
Criminal Legal Problem  & 0.842 & 0.744 & 0.790 & 0.857 & 0.698 & 0.769 & 0.875 & 0.651 & 0.747 & 0.727 & 0.558 & 0.632 & 0.974 & 0.860 & 0.914 \\
Eviction or Loss of Home & 0.909 & 0.698 & 0.789 & 0.881 & 0.860 & 0.871 & 0.800 & 0.744 & 0.771 & 0.667 & 0.791 & 0.723 & 0.971 & 0.767 & 0.857 \\
Exposure to Disaster & 0.500 & 0.029 & 0.054 & 0.727 & 0.229 & 0.348 & 0.429 & 0.086 & 0.143 & 0.471 & 0.229 & 0.308 & 0.947 & 0.514 & 0.667 \\
Family Relationship Problem & 0.773 & 0.500 & 0.607 & 0.580 & 0.853 & 0.690 & 0.871 & 0.794 & 0.831 & 0.775 & 0.912 & 0.838 & 1.000 & 0.853 & 0.921 \\
Financial Problem & 0.917 & 0.478 & 0.629 & 0.611 & 0.957 & 0.746 & 0.792 & 0.826 & 0.809 & 0.700 & 0.913 & 0.792 & 0.923 & 0.522 & 0.667 \\
Intimate Partner Problem & 0.875 & 0.824 & 0.848 & 0.620 & 0.912 & 0.738 & 0.698 & 0.882 & 0.779 & 0.674 & 0.853 & 0.753 & 1.000 & 0.912 & 0.954 \\
Job Problem & 0.950 & 0.679 & 0.792 & 0.700 & 1.000 & 0.824 & 0.792 & 0.679 & 0.731 & 0.636 & 0.750 & 0.689 & 1.000 & 0.571 & 0.727 \\
Mental Health Problem & 0.900 & 0.360 & 0.514 & 0.481 & 1.000 & 0.649 & 0.611 & 0.880 & 0.721 & 0.511 & 0.920 & 0.657 & 0.818 & 0.720 & 0.766 \\
Other Addiction Problem & 0.000 & 0.000 & 0.000 & 1.000 & 0.667 & 0.800 & 0.000 & 0.000 & 0.000 & 1.000 & 0.667 & 0.800 & 1.000 & 0.333 & 0.500 \\
Other Relationship Problem & 0.769 & 0.294 & 0.426 & 0.600 & 0.794 & 0.684 & 0.714 & 0.294 & 0.417 & 0.588 & 0.294 & 0.392 & 1.000 & 0.471 & 0.640 \\
Other Substance Abuse & 1.000 & 0.243 & 0.391 & 0.679 & 0.973 & 0.800 & 0.771 & 0.730 & 0.750 & 0.750 & 0.892 & 0.815 & 0.917 & 0.595 & 0.721 \\
Physical Health Problem & 0.864 & 0.704 & 0.776 & 0.742 & 0.852 & 0.793 & 0.833 & 0.741 & 0.784 & 0.682 & 0.556 & 0.612 & 0.952 & 0.741 & 0.833 \\
Recent Suicide of Friends or Family & 0.800 & 0.222 & 0.348 & 0.419 & 1.000 & 0.590 & 0.579 & 0.611 & 0.595 & 0.409 & 1.000 & 0.581 & 1.000 & 0.222 & 0.364 \\
School Problem & 0.882 & 0.556 & 0.682 & 0.913 & 0.778 & 0.840 & 1.000 & 0.370 & 0.541 & 0.800 & 0.741 & 0.769 & 1.000 & 0.519 & 0.683 \\
Suicide Disclosure & - & - & - & - & - & - & - & - & - & - & - & - & - & - & - \\
\bottomrule
\end{tabular}
}
\end{table}

\newpage

\begin{table}[!hbpt]
\caption{Detailed accuracy comparisons between the context retrieval in stage 1 and the relevance verification in stage 2.}
\label{tab:detailed-accuracy-comparisons}
\centering

\begin{tabular}{lcccc}
\toprule
Factor & \makecell[c]{FLAN-T5\\base} & \makecell[c]{Multi-stage\\Stage 1} & \makecell[c]{Multi-stage\\Stage 2} & \makecell[c]{Fine-Tuned\\FLAN-T5 base} \\ 
\midrule
Alcohol Problem  & 0.692 & 0.750   & 0.875   & 1.000       \\
Civil Legal Problem  & 0.586 & 0.429   & 0.667   & 0.893       \\
Criminal Legal Problem  & 0.450 & 0.429   & 0.571   & 1.000       \\
Eviction or Loss of Home & 0.778 & 0.773   & 0.941   & 0.864       \\
Exposure to Disaster & 0.429 & 0.524   & 0.545   & 0.905       \\
Family Relationship Problem & 0.632 & 0.481   & 0.625   & 0.741       \\
Financial Problem & 0.783 & 0.680   & 0.810   & 0.920       \\
Intimate Partner Problem & 0.429 & 0.522   & 0.550   & 0.609       \\
Job Problem & 0.357 & 0.684   & 0.684   & 0.895       \\
Mental Health Problem & 0.571        & 0.821   & 0.844   & 0.897       \\
Other Addiction Problem & 0.714        & 0.474   & 0.818   & 0.789       \\
Other Relationship Problem & 0.567        & 0.536   & 0.611   & 0.821       \\
Other Substance Abuse & 0.520        & 0.517   & 0.867   & 0.862       \\
Physical Health Problem & 0.636        & 0.700   & 0.842   & 0.900       \\
Recent Suicide of Friends or Family & 0.259        & 0.500   & 0.588   & 0.900       \\
School Problem & 0.526        & 0.667   & 0.857   & 0.889         \\
\bottomrule
\end{tabular}
\end{table}

\end{document}